\pdfoutput=1
\documentclass[11pt]{article}

\usepackage[final]{acl}

\usepackage{times}
\usepackage{latexsym}
\usepackage{subcaption} % Add this to your preamble
\usepackage{multirow}
\usepackage{booktabs,arydshln}
\usepackage{dcolumn}
\usepackage{array}
\usepackage{tabularx}
\usepackage{amsmath}
\usepackage{cleveref}
\usepackage{tcolorbox}
\crefname{subsection}{subsection}{subsections}

\usepackage[T1]{fontenc}
\usepackage[utf8]{inputenc}

\usepackage{microtype}

\usepackage{inconsolata}

\usepackage{graphicx}
\usepackage{todonotes}
\usepackage{booktabs}

\usepackage{cleveref}

\title{Optimizing Alignment with Less: Leveraging Data Augmentation for Personalized Evaluation} % Human-Like
\author{Javad Seraj \\ Mohammad Mahdi Mohajeri \\ Mohammad Javad Dousti \\ Majid Nili Ahmadabadi \\
Department of Electrical and Computer Engineering \\ University of Tehran \\ 
\texttt{\{javad.seraj,mehdimohajeri,mjdousti,mnili\}@ut.ac.ir}}

\usepackage{tabularx}
\begin{document}
\maketitle
\begin{abstract}

\end{abstract}
Automatic evaluation by large language models (LLMs) is a prominent topic today; however, judgment and evaluation tasks are often subjective and influenced by various factors, making adaptation challenging. 
While many studies demonstrate the capabilities of state-of-the-art proprietary LLMs in comparison to human evaluators, they often struggle to adapt to reference evaluators over time, a requirement for achieving personalized judgment.
Additionally, numerous works have attempted to apply open LLMs as judges or evaluators, but these efforts frequently overlook the limitations of working with scarce data. Personalized judgment is inherently associated with limited data scenarios, which are common in many real-world problems.
Our work aims to present a data augmentation technique to select a more effective sample from limited data in order to align an open LLM with human preference.
Our work achieves approximately 7\% improvements in Pearson correlation with a reference judge over the baseline,and  30\% improvement over the base model (Llama3.1-8B-Instruct) in the mathematical reasoning evaluation task. demonstrating that augmenting selecting more effective preference data enables our approach to surpass baseline methods.

\section{Introduction}
%Judgment is the ability to make informed decisions or reach sensible conclusions.
%
Human evaluation process is inherently subjective and heavily influenced by the evaluator's mood, which can change drastically over time.
For instance, the assessment of students' papers may vary from semester to semester, reflecting variations in the teacher's perspective or circumstances. 
This subjectivity is crucial to consider when attempting to model or replicate an evaluator's behavior. 
Additionally, automatic evaluation tasks often face data limitations, with only a small amount of feedback and scores typically available.
This highlights the importance of exploring data-efficient or effective training methods for assessment in limited-data scenarios. 
Our work presents an effective way to align an open large language model (LLM) with a reference evaluator in a data-limited setting, emphasizing on the personalized judgment across various tasks such as math and general truthful question-answering.

LLM-based evaluation has emerged as a scalable and cost-effective paradigm for assessing LM-generated text~\cite{zeng2024evaluatinglargelanguagemodels, zheng2023judgingllmasajudgemtbenchchatbot, gao2024llmbasednlgevaluationcurrent, li2024leveraginglargelanguagemodels} or human-generated text~\cite{latif2023finetuningchatgptautomaticscoring, latif2024knowledgedistillationllmautomatic, fang2024usinggpt4augmentunbalanced}.
LLMs are prompted to provide feedback along with a scalar score, which is an indicator of quality, commonly referred to as direct assessment.

Prior works employing proprietary LLMs as assessors have shown not only high correlations with human judgements but also increased speed and cost-effectiveness~\cite{10.1145/3649217.3653612, latif2023finetuningchatgptautomaticscoring, zhu2023judgelmfinetunedlargelanguage}.
These models perform well in \textit{static judgement}, where scoring is based on fixed and rigid criteria which remain unchanged over time.
However, it is not easy to personalize their behavior to follow a certain evaluator preference or policy.
Additionally, these models lack \textit{dynamic judgement}.
Dynamic judgment refers to an evaluator's ability to learn from a few samples provided by a reference judge and adjust the evaluation policy over time.
This behavior is essential to achieve personalized evaluation.

Our work presents an effective way to align an open LLM with a reference judge in a data-limited setting.
The goal is to align LLM judgment to that of the human judge.
We propose a data augmentation method in the domain of direct assessment. 
Using the proposed data augmentation and selection techniques, we achieved approximately 9\% and 7\% higher Pearson correlation compared to the baseline in math and general truthful question-answering evaluations, respectively. Additionally, we demonstrated that by selecting more effective data, our approach can outperform baseline methods.

Our contributions are summarized as follows:
\begin{itemize}
    \item Proposed a methodology to simplify dynamic judgement, tackling a previously unresolved challenge for open LLMs.
    \item Introduced a method to augment data aimed at enhancing the reasoning ability of the judge model using the chain of thought~\cite{wei2023chainofthoughtpromptingelicitsreasoning} (CoT) approach within a limited-data framework.
    \item Introduced a methodology to select highly effective instances generated from reference judgments, focusing on reducing bias in the aligned judge.
\end{itemize}

The rest of this paper is organized as explained next.
\Cref{sec:related_works} provides an overview of related work concerning LLM alignment and their judgment capabilities.
The proposed methodology is presented in \Cref{sec:method}, followed by the experimental setup and alignment procedures detailed in \Cref{sec:setup}.
In \Cref{sec:results}, we discuss the performance of our method, comparing it with existing our methods with basis-line methods. 

% \documentclass{article}
% \usepackage{makecell}

% \begin{document}
% \begin{table}[t]
% \centering
% \small
% \begin{tabular}{l c c c}
% \hline
% \textbf{Name}   & \makecell{\textbf{Training} \\ \textbf{Samples}} & \makecell{\textbf{Fine/Coarse} \\ \textbf{Grained}} & \makecell{\textbf{Human/Machine} \\ \textbf{Response}} \\ \hline
% \multicolumn{4}{l}{\textbf{Open LLMs}} \\ \hline
% PandaLM          &            &                     &                     \\ 
% JudgeLM          &            &                     &                     \\ 
% Prometheus       &            &                     &                     \\ \hline
% \multicolumn{4}{l}{\textbf{Closed LLMs}} \\ \hline
% Model X          &            &                     &                     \\ 
% Model Y          &            &                     &                     \\ \hline
% \end{tabular}
% \caption{Comparison of Open and Closed LLMs based on evaluation.}
% \label{tab:llm_comparison}
% \end{table}

\section{Related Works}

\begin{figure*}[!htb]
    \centering
    % First two images side by side
    \begin{subfigure}{0.35\textwidth}
        \centering
        \includegraphics[width=\textwidth]{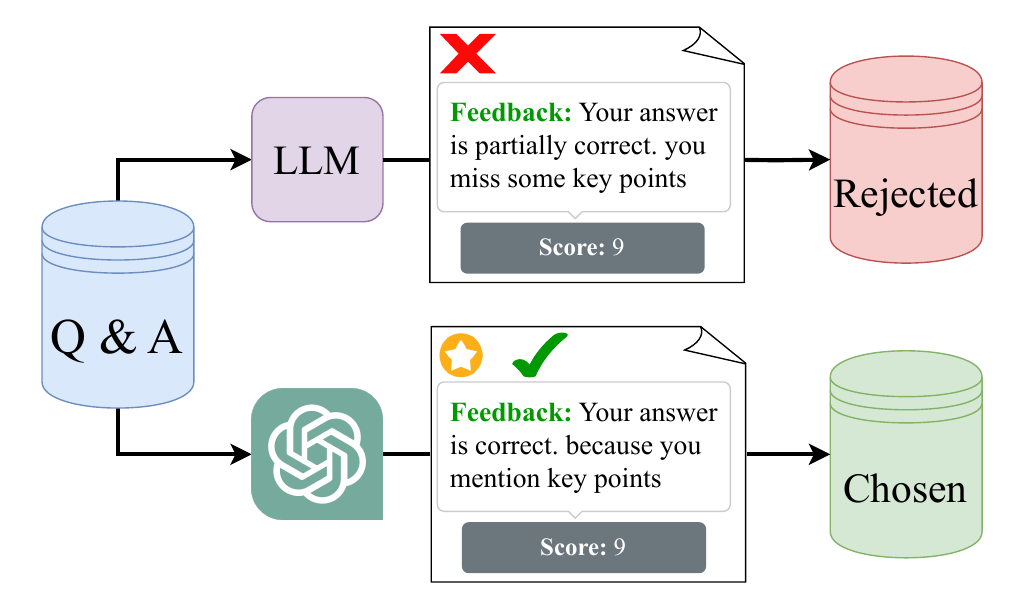}
        \caption{Na\"ive Data Creation}
        \label{fig:classic_method}
    \end{subfigure}
    \hfill
    \begin{subfigure}{0.64\textwidth}
        \centering
        \includegraphics[width=\textwidth]{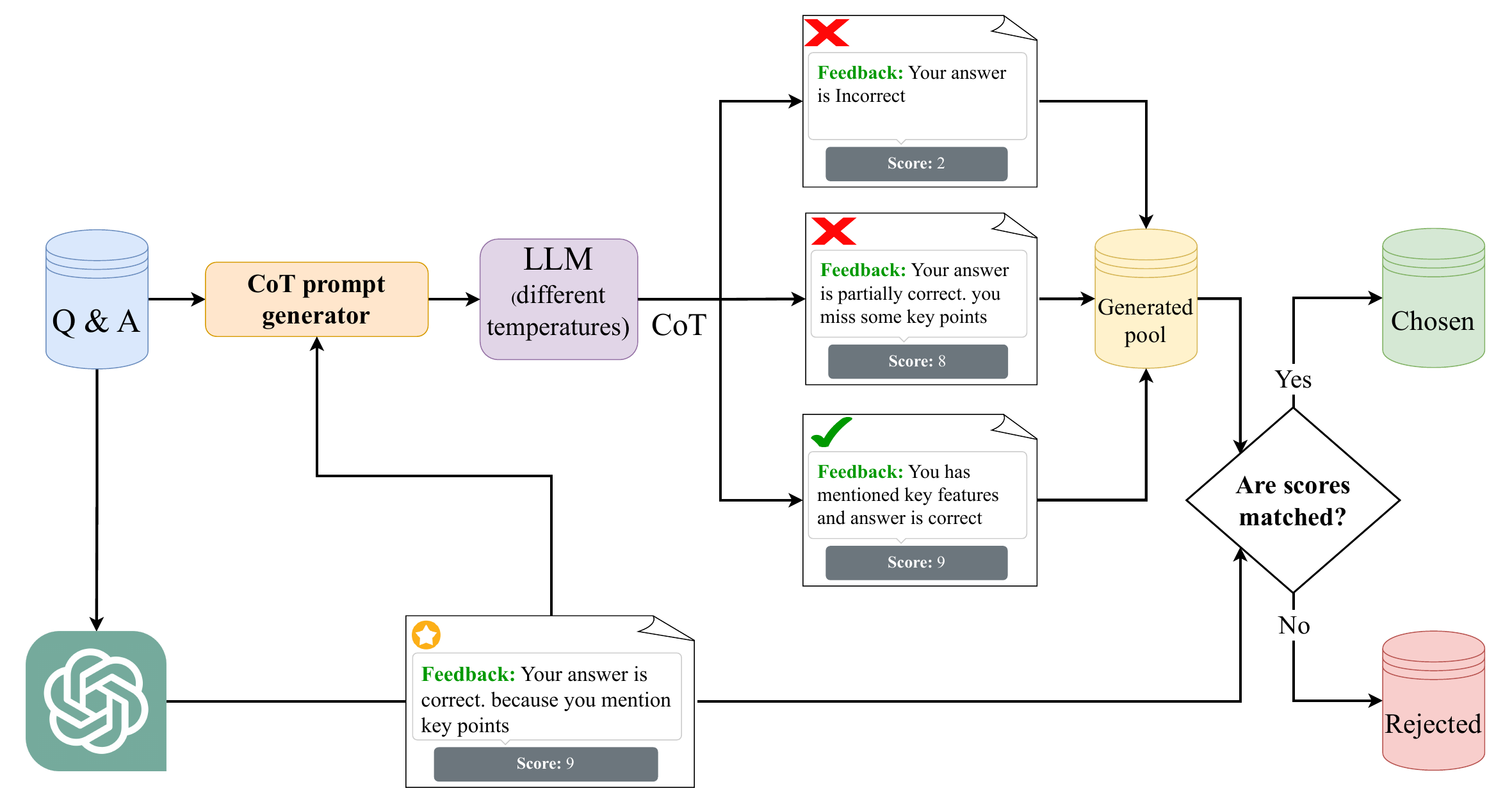}
        \caption{Pool of Feedback}
        \label{fig:pool_of_feedback}
    \end{subfigure}
    
    % Third image below spanning both columns
    \begin{subfigure}{0.95\textwidth}
        \centering
        \includegraphics[width=\textwidth]{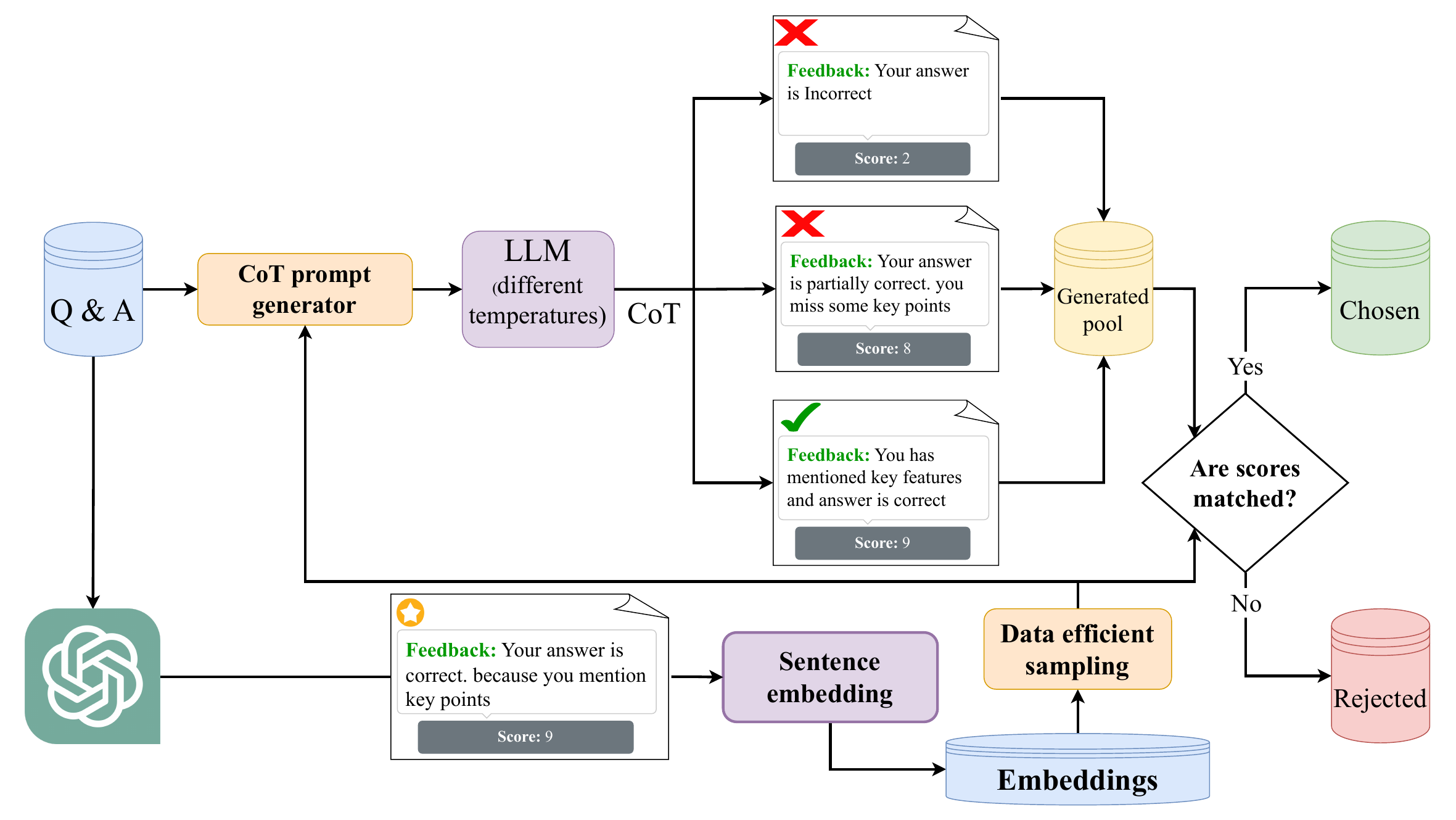}
        \caption{Efficient Sampling}
        \label{fig:efficient_sampling}
    \end{subfigure}
    \caption{Different approaches for preference data creation. (a) The na\"ive data creation approach, (b) Pool of feedback approach, and (c) The efficient sampling approach.}
%    \caption{Different approaches for preference data creation. (a) The na\"ive data creation approach considers reference feedback as gold and chosen feedback and generated feedback by selected LLM as rejected feedback. (b) Our pool of feedback approach considers reference feedback as gold. This approach generates a CoT prompt based on gold feedback and respective question and answer. Used LLM run with different temperatures and generates diverse feedback By using CoT. Feedback with the same score to gold are considered as chosen and feedback with the different score to gold are considered as rejected. (c) The efficient sampling approach is similar to the pool of feedback with differences in using gold feedback. This approach selects effective samples from gold feedback based on their embedding similarity and uses selected feedback instead of all feedback.}
    \label{fig:three_figures}
\end{figure*}

\label{sec:related_works}
\subsection{LLM as a Judge}
Recently, employing LLMs as judges or evaluators has gained attention as a reliable and accurate paradigm to mimic the accuracy that human evaluation offers. 
Many works use proprietary LMs such as GPT-4 as reference judges. These models have shown high agreement with human assessments and text evaluations in both direct assessment and relevance judgments~\cite{zheng2023judgingllmasajudgemtbenchchatbot, latif2023finetuningchatgptautomaticscoring}. 

Kim et al.\cite{kim2024prometheusinducingfinegrainedevaluation} and Zhu et al.~\cite{zhu2023judgelmfinetunedlargelanguage} introduced methods for generating feedback datasets for both pairwise judgments and direct scoring assessment.
These models used fine-tuned Llama model to align them against GPT-4, where proven high correlation. 
Kim et al.~\cite{kim2024prometheus2opensource} fine-tuned two models: the first model assigns a score to LLM responses using a direct assessment feedback dataset, and the second model is fine-tuned for pairwise ranking using pairwise feedback. Merging techniques were then employed to create a unified model that performed well in both pairwise ranking and direct scoring tasks.
\subsection{Human Preference Alignment}
LLMs excel at text generation but face challenges in instruction following and alignment with human expectations, spurring research to address these limitations~\cite{wang2023aligning}.
Supervised Fine-Tuning (SFT) has emerged as a key method for aligning LLMs with human expectations, demonstrated by its effectiveness in InstructGPT~\cite{ouyang2022instructGPT} and subsequent widespread adoption in alignment research~\cite{wang2023aligning}.

InstructGPT incorporates Reinforcement Learning (RL) after its SFT step.
This approach builds on the concept of reinforcement learning from human feedback (RLHF)~\cite{ouyang2022instructGPT, rlhf}.
Following InstructGPT, several methods such as DPO~\cite{rafailov2024dpo}, RAFT~\cite{dongraft}, and RLAIF~\cite{lee2023rlaif} have introduced their work based on the RLHF method.
They have enhanced their techniques to improve alignment or data efficiency.

DPO has introduced an approach to RLHF which simplifies the process through closed-form policy extraction and a straightforward classification loss~\cite{rafailov2024dpo}.
DPO has gained significant traction in the field of LLM alignment, with prominent open models such as Llama3~\cite{dubey2024llama3} and Mixtral~\cite{jiang2024mixtral} adopting this methodology for their alignment strategies.

\subsection{Data Efficient Alignment}
The size and quality of data impact the cost and time of LLMs training~\cite{kaddour2023challenges, kaplan2020scaling}.
Efficient use of data can reduce the number of iterations needed for training~\cite{kaddour2023challenges}.
Some works focus on improving training data quality by synthesizing high-quality data and filtering low-quality data.

Phi3 combines filtered web content and synthetic LLM-generated data from diverse sources to improve training data quality~\cite{abdin2024phi3}.
Llama3 emphasizes Llama2's capability to identify high-quality data and uses Llama2 to build a text classifier to improve training data quality~\cite{dubey2024llama3}.
FineWeb-Edu improved FineWeb by creating an educational quality classifier trained on LLama3-70B-Instruct annotations, using it to select only the most educational web pages.
FineWeb-Edu, demonstrates superior performance on common benchmarks, highlighting the effectiveness of classifiers trained on synthetic data~\cite{penedo2024fineweb}.

Evans et al. argue that selecting data batches together is more effective than choosing examples independently~\cite{evans2024batch_data_selection}.
This work uses dependencies between data points, allowing for better measurement of how learnable a batch is as a whole.
Some other works focus on developing data-efficient algorithms for model training and alignment.
The KTO method presents an alignment approach grounded in prospect theory, utilizing only binary preference signals rather than paired comparisons.
This innovation allows for more efficient use of alignment data, effectively doubling the data points available compared to previous methods~\cite{ethayarajh2024kto}.

% Second version of table, with booktabs.

% \begin{table*}[ht]
% \centering
% \begin{tabular}{lcccccl}\toprule
% & \multicolumn{3}{c}{$\tol=\tols$} & \multicolumn{3}{c}{$\tol=\told$}
% \\\cmidrule(lr){2-4}\cmidrule(lr){5-7}
%            & $mv$  & Rel.~err & Time    & $mv$  & Rel.~err & Time\\\midrule
% \trigmv    & 11034 & 1.3e-7 & 3.9 & 15846 & 2.7e-11 & 5.6 \\
% \trigexpmv & 21952 & 1.3e-7 & 6.2 & 31516 & 2.7e-11 & 8.8 \\
% \trigblock & 15883 & 5.2e-8 & 7.1 & 32023 & 1.1e-11 & 14.0\\
% \expleja   & 11180 & 8.0e-9 & 4.3 & 17348 & 1.5e-11 & 6.6 \\\bottomrule
% \end{tabular}
% \caption{Performance comparison for different methods with two tolerance levels $\tol=\tols$ and $\tol=\told$.}
% \end{table*}

% \begin{table*}[ht]
% \centering
% \begin{tabular}{lccc}\toprule
%        & Fine-tuning Method & Number of Training Dataset & Fine/Coarse \\ \midrule
% \ JudgeLM~\cite{zhu2023judgelmfinetunedlargelanguage}   &     SFT  &  110k & coarse \\
% \ Prometheus~\cite{kim2024prometheusinducingfinegrainedevaluation} &     SFT  &  k & fine \\
% Prometheus2\cite{kim2024prometheus2opensource}  &     SFT  & K & fine \\
%\cite{wang2024directjudgementpreferenceoptimization}  & SFT+DPO  &  k  &  fine\\\bottomrule
% \end{tabular}
% \caption{Comparison of models, their training datasets, and methods used.}
% \end{table*}

\section{Data-Efficient Judgement}
\label{sec:method}

In this section, we present a data-efficient approach for aligning an LLM with a reference judge. 
While our work primarily focuses on machine-generated text, it can also be extended to human-generated text.

First, we introduce the dataset format.
Each dataset includes questions and corresponding responses as input, with feedback and scores as output.
Next, we outline our methodology for augmenting the dataset, followed by a systematic approach to selecting the most effective judgment instances.
Finally, we detail our method for training a data-efficient evaluator.

\subsection{Data Curation and Augmentation}
Many studies build a dataset for the judgment task~\cite{kim2024prometheus2opensource, zhu2023judgelmfinetunedlargelanguage, kim2024biggenbenchprincipledbenchmark}. 
Assessment tasks require strong reasoning ability because reasoning helps the judge make more accurate and fair decisions~\cite{wang2024directjudgementpreferenceoptimization}.

we show in table~\Cref{tab:pearson_results} open LLMs, such as Llama-3.1-8B-Instruct are ineffective evaluators. 
Previous studies~\cite{kim2024prometheusinducingfinegrainedevaluation, zhu2023judgelmfinetunedlargelanguage} have demonstrated that LLMs such as Llama2 (7B, 13B, 34B) and Mistral 7B show low agreement with human evaluators or proprietary LLMs like GPT-4.
In the rest of this section, we define our seed dataset and then introduce our methodology to augment data using a CoT approach.
Aside from the benefit of growing our dataset, CoT enhances the model's reasoning ability which further improves the accuracy and fairness of its judgment calls~\cite{mukherjee2023orcaprogressivelearningcomplex}.

\subsubsection{Seed for Preference Dataset}

We start with a question and response dataset.
The response can come from either a human or an LLM depending on the actual task of interest.
For each response, we collect a feedback and score (within a prespecified range) from a reference judge.
The aim is to use this \textit{seed dataset} to improve a base LLM judgement performance, i.e., making it behave similar to a reference judge.

To achieve this goal, we try to align the base LLM, which requires preference data (chosen and rejected) feedbacks and scores.
More precisely, the training dataset comprises of $n$ samples, with each sample represented as a tuple containing the following elements:

\begin{equation}
\label{eq:data}
    \mathcal{D} = \Big\{(q_i, r_i, f_i^g, F_i^r, F_i^c)\Big\},
\end{equation}
\noindent where $q_i$ is a question, $r_i$ denotes a response to $q_i$, $f_i^g$ indicates the golden feedback and score for $r_i$, which is generated by a reference judge, $F_i^r$ is a list of rejected feedbacks and their respective scores related to $r_i$, and $F_i^c$ is a list of chosen feedbacks and their respective scores related to $r_i$.
Note that $q_i$, $r_i$, and $f_i^g$ are taken from a seed dataset, whereas 
$F_i^r$ and $F_i^c$ are provided by the base LLM.
The following subsections explains how $F_i^r$ and $F_i^c$ are generated.

% \begin{figure}[htbp]
%     \centering
%     \includegraphics[width=\textwidth]{data efficient judgement-classic method.drawio.png} % Adjust the width to fit the column width
%     \caption{Your caption here}
%     \label{fig:classic_method}
% \end{figure}

% \begin{figure}[htbp]
%     \centering
%     \includegraphics[width=\textwidth]{data efficient judgement-Pool of feedback method.drawio.png} % Adjust the width to fit the column width
%     \caption{Your caption here}
%     \label{fig:pool_of_feedback}
% \end{figure}

% \begin{figure*}[htbp]
%     \centering
%     \includegraphics[width=\textwidth]{data efficient judgement-efficient sampling method.drawio.png} % Adjust the width to fit the column width
%     \caption{Your caption here}
%     \label{fig:efficient_sampling}
% \end{figure*}

\subsubsection{Na\"ive Data Creation Approach} 
\label{sec:simple_aproach}

The base LLM is required to generate feedback and a corresponding score within a prespecified range.
In this approach, we take all feedbacks and scores in $F^g_i$ as chosen feedbacks and place them in $F^c_i$.
Next, we apply the base LLM to generate a feedback and a score for each response, i.e., $r_i$.
These feedbacks and scores construct the rejected list, i.e., $F^r_i$, which we assume have inferior quality compared to those come from the reference judge.
$F^c_i$ and $F^r_i$ form the basis of our preference data for aligning the base LLM to the reference judge (See \Cref{fig:classic_method}).

\subsubsection{Pool of Feedback Approach}
\label{sec:pool_of_feedback_approach}
As shown in \Cref{fig:pool_of_feedback}, in this approach, we use the base LLM to generate multiple feedback and score pairs for each response.
The base LLM is provided with the reference judge's feedback as auxiliary information, treating it as a hint.
This allows the base LLM to leverage its own reasoning abilities to generate better feedback and corresponding score using CoT.

To diversify employed chains of thought, the base LLM uses a range of temperatures from $0.2$ to $1.4$. In this way, for each $(q_i, r_i)$ pair, the base LLM uses different reasoning to evaluate the response. 
These generated feedbacks, produced at various temperatures is divided into two parts:
\begin{enumerate}
    \item \textbf{Chosen pool}: Chains of thought which led to correct scores, i.e., where the score matches the one assigned by the reference judge.
    \item \textbf{Rejected pool}: Chains of thought which led to incorrect scores, i.e., where the score does not match the one assigned by the reference judge.
\end{enumerate}

We use pairs of chosen and rejected feedback from these two pools to create $F_i^r$ and $F_i^c$.
Note that, unlike the Na\"ive data generation approach, here we do not use reference judgments directly.

\begin{figure}[t]
    \centering
    \begin{subfigure}{\columnwidth}
        \includegraphics[width=\columnwidth, trim={0 18 0 0},clip]{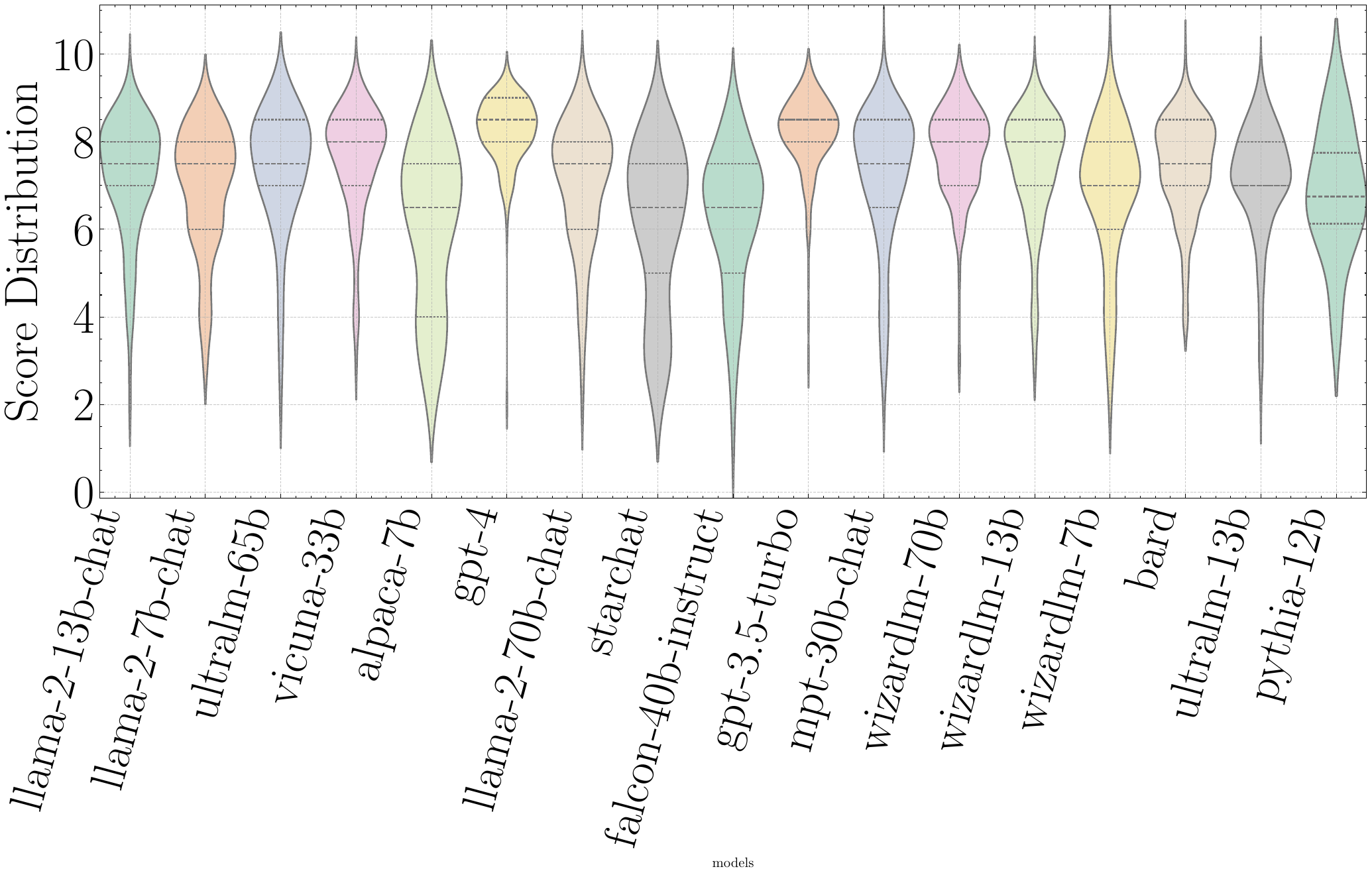} % Adjust the width to fit the column width
        \caption{TruthfulQA}
\label{fig:truthful_score_dist}
    \end{subfigure}
       \begin{subfigure}{\columnwidth}
        \includegraphics[width=\columnwidth,trim={0 18 0 0},clip]{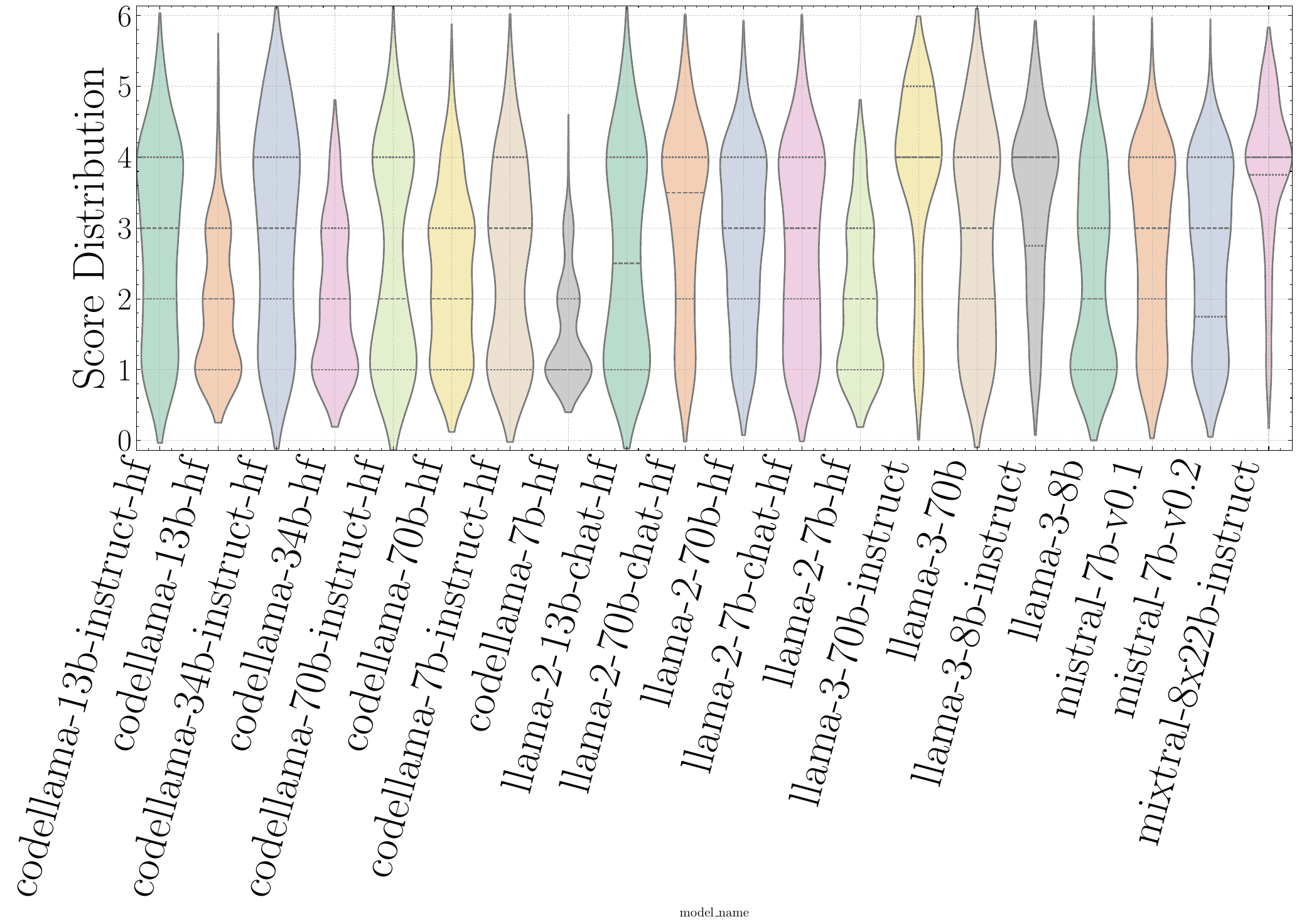} % Adjust the width to fit the column width
        \caption{BigGen-Bench}
        \label{fig:bgb_score_dist}
    \end{subfigure}
    \caption{Scores distribution for different LLMs in (a) BigGen-Bench and (b) TruthfulQA datasets. Note that the distribution of scores on BigGen-Bench is discrete, whereas the scores on TruthfulQA is continuous.}
    \label{fig:score_dist}
\end{figure}

\subsubsection{Efficient Sampling Approach}
\label{sec:efficient_sampling}

In this section, we introduce a methodology for selecting more effective samples from the reference judge.
This approach helps to align the model more accurately with the reference judge, reducing the risk of biases that can emerge during the alignment process.

As shown in \Cref{fig:efficient_sampling}, instead of using the entire set of golden feedback, we choose a subset of it.
First, the sentence embedding is calculated for each golden feedback, i.e., $f_i^g$.
Next, these embeddings are clustered based on their similarity.
We use the K-means algorithm to perform the clustering.
Within each cluster, feedback are grouped based on their respective score.
Finally, a representative batch of feedback is selected from each group in every cluster.
The batch size is determined such that the overall number of feedback for each score becomes balanced across various scores.
This approach addresses the imbalance in the distribution of scores within the seed dataset, ensuring that the trained model does not become biased toward certain scores.

\begin{figure*}[!htb]
    \centering
    % First two images side by side
    \begin{subfigure}{0.49\textwidth}
        \centering
        \includegraphics[width=\textwidth]{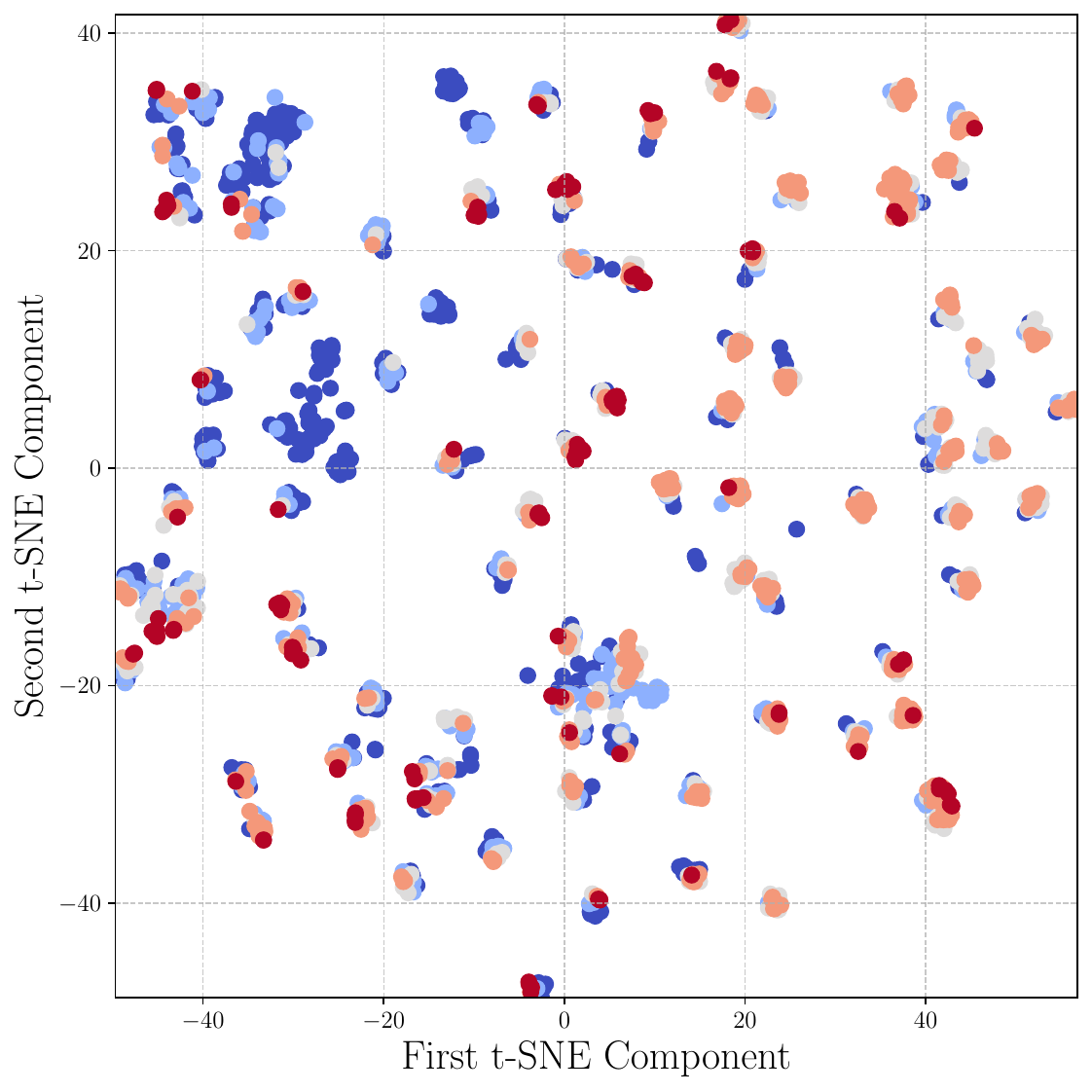}
        \caption{All Data}
        \label{fig:bgb_all_points}
    \end{subfigure}
    \hfill
    \begin{subfigure}{0.49\textwidth}
        \centering
        \includegraphics[width=\textwidth]{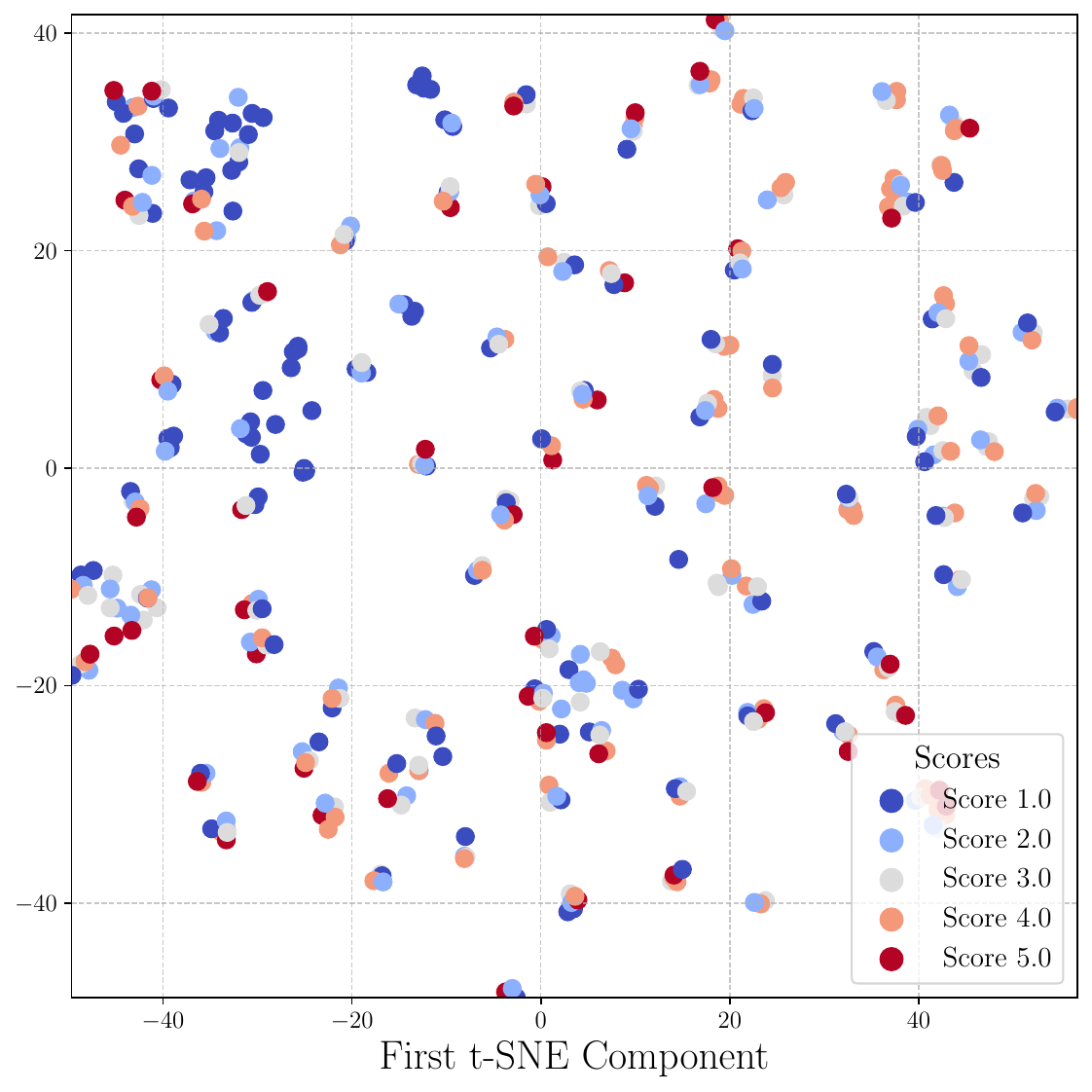}
        \caption{Sampled Data}
        \label{fig:bgb_selected_points}
    \end{subfigure}
    \caption{Selected data from BigGen-Bench benchmark, where generated feedback and respective scores are discrete. (a) The distribution of the entire dataset embeddings and (b) the distribution of selected samples using the efficient sampling approach. Note that both size is different to balance data across various scores.}
    %Previous approaches select all feedback from the dataset for data augmentation. Our efficient sampling approach clusters feedback based on their similarity. This approach groups data in every cluster based on respective scores and selects a representative batch of feedback from each group. Both size is different to balance data across various scores (b).}
    \label{fig:bgb_points}
\end{figure*}

\begin{figure*}[!htb]
    \centering
    % First two images side by side
    \begin{subfigure}{0.49\textwidth}
        \centering
        \includegraphics[width=\textwidth]{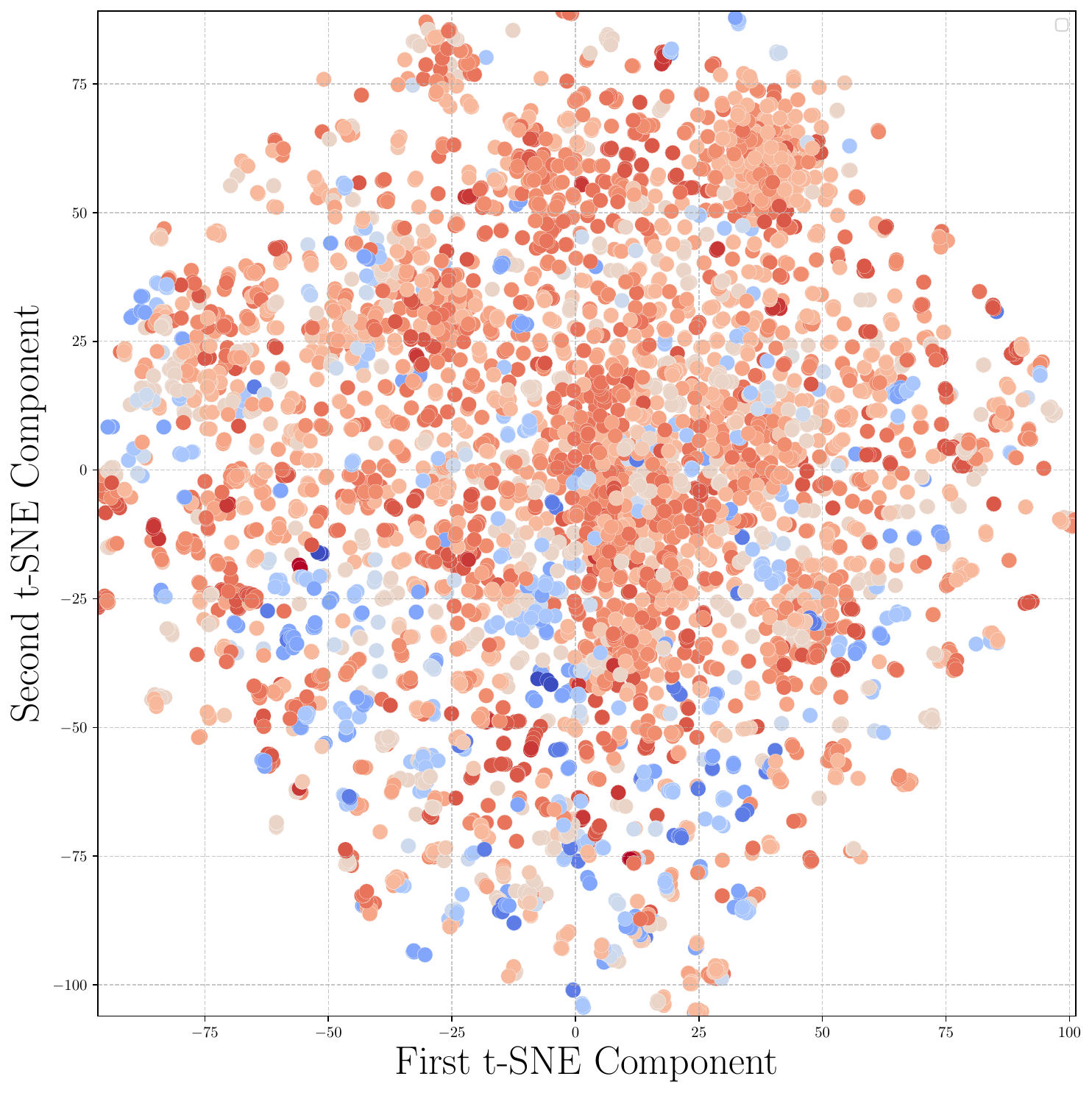}
        \caption{All Data}
        \label{fig:truthful_all_points}
    \end{subfigure}
    \hfill
    \begin{subfigure}{0.49\textwidth}
        \centering
        \includegraphics[width=\textwidth]{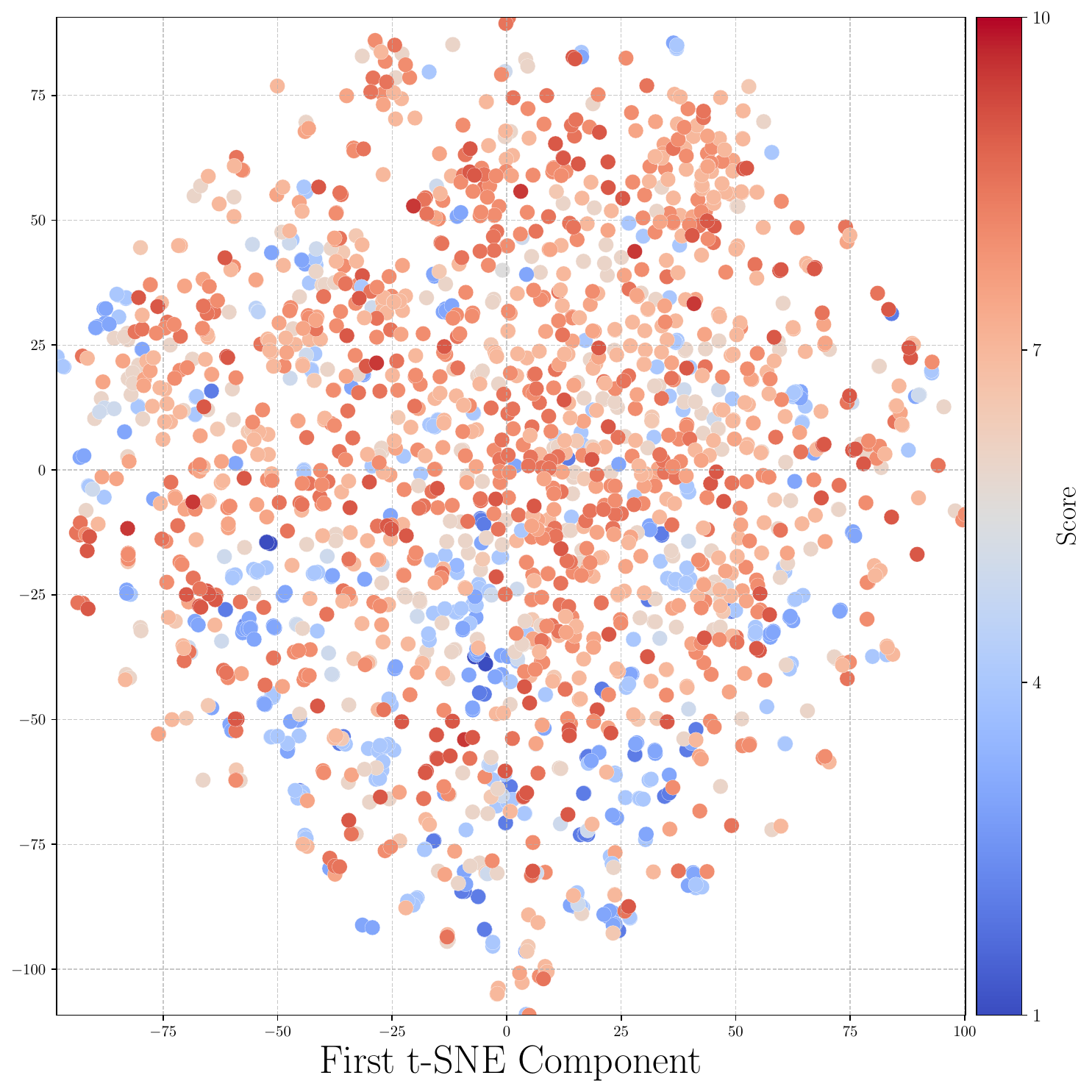}
        \caption{Sampled Data}
        \label{fig:truthfulQA_selected_points}
    \end{subfigure}
    
    \caption{Selected data from TruthfulQA benchmark, where generated feedback and respective scores are continuous. (a) The distribution of the entire dataset embeddings and (b) the distribution of selected samples using the efficient sampling approach. Note that both size is different to balance data across various scores.}
    %{Previous approaches select all feedback from the dataset for data augmentation (a). Our efficient sampling approach clusters feedback based on their similarity. This approach groups data in every cluster based on the range of scores and selects a representative batch of feedback from each group. Bath size is different to balance data across various scores (b).}
    \label{fig:truthful_points}
\end{figure*}

Similar to the previous subsection, each chosen feedback ($f_i^g$), along with its associated question ($q_i$) and response ($r_i$) to it are used for chosen and rejected feedback generation $(F_i^r, F_i^c)$.
\Cref{fig:bgb_points} shows the embeddings of all feedback and sampled feedback for BigGen-Bench.
\Cref{fig:truthful_points} shows all feedback and sampled feedback for TruthfulQA.
As shown, the diversity of data is preserved in sampled data.
Because data is sampled from different clusters.
Also, sampled feedback are more balanced based on scores.
Because sampling ratios are different for different scores.

\section{Experiment Setup}
\label{sec:setup}
\begin{table}[t]
    \centering
    \resizebox{\columnwidth}{!}{
    \begin{tabular}{lcccc}
        \toprule
        & \textbf{BigGen-Bench} & \textbf{TruthfulQA} \\
        \cmidrule{1-3} 
        
        Seed for preference dataset & 2 k& 2.4 k   \\
        Na\"ive & 2 k& 2.4 k  \\
        Pool of feedback & 4.9 k& 6.7 k  \\
        Efficient sampling & 2.3 k& 2.2 k  \\

        \bottomrule
    \end{tabular}
}
    \caption{Size of created data with different approaches}
    \label{tab:traiinig_sample_size}
\end{table}
\subsection{Alignment Dataset}

We extract specific partitions for our analysis from two feedback datasets: UltraFeedback~\cite{cui2024ultrafeedbackboostinglanguagemodels} and BigGen-Bench~\cite{kim2024biggenbenchprincipledbenchmark}. From UltraFeedback, we utilize the TruthfulQA~\cite{lin-etal-2022-truthfulqa} partition, a benchmark designed to assess whether language models provide truthful responses to questions across diverse categories such as health, law, finance, and politics.
This benchmark contains 817 questions and 3,268 responses across 38 categories, challenging models to avoid generating false answers that may stem from imitating human texts.
These questions were given to different language models, and their corresponding scores were generated by GPT-4 on a scale from 1 to 10.

The score distributions for each evaluated model's responses are shown in ~\Cref{fig:truthful_score_dist}.
As illustrated in ~\Cref{fig:truthful_score_dist}, proprietary LLMs, such as ChatGPT, tend to achieve higher scores, with a distribution skewed toward the upper range. 
Furthermore, scores in the 1 to 6 range are relatively rare across many models, suggesting a potential for biases in the alignment process.

For BigGen-Bench, we extract 2,000 samples from the reasoning partition, which evaluates the mathematical reasoning of models.
Additionally, the BigGen-Bench dataset includes detailed scoring rubrics for each question, offering a more structured evaluation framework.
BigGen-Bench is structured with model performance scores on a 5-point Likert scale~\cite{Robinson2014}, with columns representing model names and scores for different capabilities.
These scores come from multiple judge models, including GPT-4, Claude-3-Opus, and Prometheus-2 (see ~\Cref{fig:bgb_score_dist}).

\begin{table*}[ht]
    \centering
    \resizebox{\textwidth}{!}{
    \begin{tabular}{lcccccc}
        \toprule
        \multirow{2}{*}{\textbf{Model}}& \multicolumn{3}{c}{\textbf{BigGen-Bench}}& \multicolumn{3}{c}{\textbf{TruthfulQA}} \\
        \cmidrule{2-4}\cmidrule(l){5-7}
        & \textbf{Pearson} & \textbf{Spearman} & \textbf{Kendall's Tau} & \textbf{Pearson} & \textbf{Spearman} & \textbf{Kendall's Tau}\\
        \midrule
        
        GPT-4-turbo     & 0.88 & 0.86 & 0.79  &  N/A  &  N/A    & N/A  \\
        GPT-4           & 1.00  &   1.00 & 1  &  N/A  &  N/A    & N/A  \\
        Claude-3-Opus   &0.75 &0.74 &0.63  &  N/A  &  N/A    & N/A \\
        \midrule
        Llama-3.1-8B-Instruct   & 0.31&0.35  & 0.30 & 0.29& 0.30&0.26\\
        Na\"I've Approach &0.54&0.61&0.51 & 0.36 & 0.36 & 0.30\\
        Ours (Pool of Feedback) &\textbf{0.63}&\textbf{0.66}&  \textbf{0.53}  & \textbf{0.43}& 0.34 & 0.30\\
        Ours (Efficient Sampling) &0.57&0.65& 0.48&\textbf{0.43} & \textbf{0.40}& \textbf{0.35}\\
        
        \bottomrule
    \end{tabular}
}
    \caption{Pearson, Spearman, and Kendall's Tau correlations between reference judge (GPT-4) and the evaluator LLM on BigGen-Bench and TruthfulQA datasets.}
    \label{tab:pearson_results}
\end{table*}

% \begin{table*}[h]
%     \resizebox{\columnwidth}{!}{
%     \begin{tabular}{lccc}
%         \toprule
%         \multirow{2}{*}{\textbf{Model}}& \multicolumn{3}{c}{\textbf{BigGen-Bench}}\\
%         \cmidrule{2-4}
%         & \textbf{Pearson} & \textbf{Spearman} & \textbf{Kendall's Tau} \\
%         \midrule
        
%         GPT-4-turbo     & 0.88 & 0.86 & 0.79   \\
%         GPT-4           & 1.00  &   1.00 & 1    \\
%         Claude-3-Opus   &0.75 &0.74 &0.63  &   \\
%         \midrule
%         Llama-3.1-8B-Instruct   & 0.31&0.35  & 0.30 \\
%         Na\"I've Approach &0.54&0.61&0.51 \\
%         Ours (Pool of Feedback) &\textbf{0.63}&\textbf{0.66}&  \textbf{0.53} \\
%         Ours (Efficient Sampling) &0.57&0.65& 0.48\\
%         \bottomrule
%     \end{tabular}
% }
%     \caption{Pearson, Spearman, and Kendall's Tau correlations between reference judge (Human) and the evaluator LLM on BigGen-Bench dataset}
%     \label{tab:human_corr_result}
% \end{table*}

After collecting the seed dataset, we applied three techniques discussed in ~\Cref{sec:method}, to generate and augment the feedback\footnote{This dataset will be released to the public upon acceptance of the paper}.
Prompts used for augmenting the preference data (rejected and chosen pools) are shown in ~\Cref{sec:prompts}.
Afterward, the DPO algorithm was applied to align the model with the reference judge.
The specifics of this process are provided next.

\subsection{Alignment Recipe}
To align the model with the reference judge, we apply DPO for each dataset separately.
Rather than tuning the full model, we train a LoRA~\cite{hu2021loralowrankadaptationlarge} adapter, which avoids the need to keep both a reference and trained model in GPU memory.
For LoRA hyperparameters, we set $r = 32$, $\alpha = 16$, and a dropout of $0.05$, targeting all linear layers.
We use the AdamW optimizer with a peak learning rate of $5 \times 10^{-6}$, following a cosine decay schedule after 0.05 warmup steps. 
For each dataset, DPO was applied to Llama-3.1-8B-Instruct across three experiments, which we detail as follows:

\subsubsection{Alignment Experiments}
In the first experiment, DPO was applied to the pairs ($F^g_i$, $F^r_i$). 
In the second experiment, it was applied to pairs of ($F^c_i$, $F^r_i$), generated using the pool of feedback approach, as explained in ~\Cref{sec:pool_of_feedback_approach}.
In the third experiment, we used the efficient sampling method described in ~\Cref{sec:efficient_sampling} to improve the alignment process by selecting more effective samples from $F^g_i$ to form preference pairs ($F^c_i$, $F^r_i$).
This method helps mitigate biases that may arise during augmentation. 
The details of each preference dataset are provided in ~\Cref{tab:traiinig_sample_size}.

%
% \subsubsection{Alignment to a Feedback Pool}
% In the second experiment, DPO was applied to pairs of ($F^g_i$,$F^r_i)$. In the third experiment, we applied the efficient-sampling method, described in ~\Cref{sec:efficient_sampling}, to enhance the alignment process by selecting more effective samples from the dataset. This method helps reduce biases that may arise during augmentation.

% \subsubsection{Alignment with Efficient Feedback}
% In the third experiment, we applied the efficient-sampling method, described in ~\Cref{sec:efficient_sampling}, to enhance the alignment process by selecting more effective samples from the dataset. This method helps reduce biases that may arise during augmentation. For example, in the TruthfulQA dataset (see ~\Cref{fig:truthfulqa_score_dist}), in the TruthfulQA dataset, responses that are assigned a score of 9 appear significantly more often than those with a score of 1. This imbalance can create a bias during the augmentation process, as the model may be exposed to the higher-scoring responses more frequently. As a result, the model could over-prioritize these high-frequency scores, skewing its predictions and evaluations.

% Details about the training process can be found in Appendix ~\ref{sec:appendix}

\subsection{Evaluation Setup}
In this section, we describe our experimental setup for assessing evaluator LMs, a process we refer to as meta-evaluation since we are evaluating the performance of the evaluator itself. We employ Pearson, Spearman, and Kendall-Tau as performance metrics to measure scoring correlations against the reference evaluator (GPT-4). This demonstrates how well the model aligns with the reference evaluator.

\section{Results}
\label{sec:results}

In this section, we present the results of our judge model and analyze the outcomes, comparing the three approaches discussed in~\Cref{sec:method}. The Pearson correlation between the scores assigned by the aligned models and those assigned by the reference judge is shown in~\Cref{tab:pearson_results}.

In the BigGen-Bench dataset, as seen in~\Cref{fig:bgb_score_dist}, the response scores are well-balanced. The \textit{pool of feedback} approach in BigGen-Bench shows more agreement with the reference judge (GPT-4). Another insight from this table is that aligning the model with the reference judge also improves its alignment with human judgments. As observed in~\Cref{tab:pearson_results}, aligning the model with GPT-4 increases its agreement with human evaluations. We also note that when the training dataset size is reduced from 4.9k to 2.7k samples, as shown in~\Cref{tab:traiinig_sample_size}, the model's performance still surpasses that of the Na\"ive data generation approach described in~\Cref{sec:simple_aproach}. 
These results suggest that COT reasoning can help the model perform evaluation and judgment tasks more effectively.

For the TruthfulQA dataset, as illustrated in ~\Cref{fig:truthful_score_dist}, the distribution is skewed. Scores like 9 occur frequently, while lower scores, such as 1, are less common in the seed dataset. This imbalance carries over to the augmented dataset, affecting its distribution. Nonetheless, by employing an efficient sampling strategy, the skewness is better managed, allowing for more representative samples and improved performance in judgment tasks. The observed decrease in correlation between GPT-4 and the pool of feedback can be attributed to the amplified skewness in the augmented data.  

\section{Conclusion}
Although large language models (LLMs) showed promise for automatic evaluation, adapting them to subjective judgment tasks in data-scarce environments remained challenging. Proprietary LLMs often failed to maintain alignment with reference evaluators over time, and efforts to use open LLMs frequently overlooked the limitations of limited data. In this work, we introduced a data augmentation technique that enhanced the selection of effective preference samples, achieving approximately a 7\% improvement in Pearson correlation with a reference judge compared to baseline methods, and about a 30\% improvement over the base model (Llama3.1-8B-Instruct) in the mathematical reasoning evaluation task.

\section{Limitation}
While our approach demonstrates significant improvements in aligning LLM judgment with human evaluators, several limitations remain. First, our work is constrained by the availability of feedback data. Although we explored data augmentation techniques to mitigate this, the alignment performance could benefit from more diverse datasets. Additionally, our method focuses on a narrow range of tasks, such as math and truthful question-answering, which limits its generalizability to other domains requiring more judgment.

\bibliography{acl_latex}

\begin{thebibliography}{33}
\providecommand{\natexlab}[1]{#1}

\bibitem[{Abdin et~al.(2024)Abdin, Jacobs, Awan, Aneja, Awadallah, Awadalla, Bach, Bahree, Bakhtiari, Behl et~al.}]{abdin2024phi3}
Marah Abdin, Sam~Ade Jacobs, Ammar~Ahmad Awan, Jyoti Aneja, Ahmed Awadallah, Hany Awadalla, Nguyen Bach, Amit Bahree, Arash Bakhtiari, Harkirat Behl, et~al. 2024.
\newblock Phi-3 technical report: A highly capable language model locally on your phone.
\newblock \emph{arXiv preprint arXiv:2404.14219}.

\bibitem[{Casper et~al.(2023)Casper, Davies, Shi, Gilbert, Scheurer, Rando, Freedman, Korbak, Lindner, Freire et~al.}]{rlhf}
Stephen Casper, Xander Davies, Claudia Shi, Thomas~Krendl Gilbert, J{\'e}r{\'e}my Scheurer, Javier Rando, Rachel Freedman, Tomasz Korbak, David Lindner, Pedro Freire, et~al. 2023.
\newblock Open problems and fundamental limitations of reinforcement learning from human feedback.
\newblock \emph{arXiv preprint arXiv:2307.15217}.

\bibitem[{Cui et~al.(2024)Cui, Yuan, Ding, Yao, He, Zhu, Ni, Xie, Xie, Lin, Liu, and Sun}]{cui2024ultrafeedbackboostinglanguagemodels}
Ganqu Cui, Lifan Yuan, Ning Ding, Guanming Yao, Bingxiang He, Wei Zhu, Yuan Ni, Guotong Xie, Ruobing Xie, Yankai Lin, Zhiyuan Liu, and Maosong Sun. 2024.
\newblock \href {https://arxiv.org/abs/2310.01377} {{UltraFeedback}: Boosting language models with scaled {AI} feedback}.
\newblock \emph{Preprint}, arXiv:2310.01377.

\bibitem[{Dong et~al.(2023)Dong, Xiong, Goyal, Zhang, Chow, Pan, Diao, Zhang, Shum, and Zhang}]{dongraft}
Hanze Dong, Wei Xiong, Deepanshu Goyal, Yihan Zhang, Winnie Chow, Rui Pan, Shizhe Diao, Jipeng Zhang, Kashun Shum, and Tong Zhang. 2023.
\newblock Raft: Reward ranked finetuning for generative foundation model alignment.
\newblock \emph{arXiv preprint arXiv:2304.06767}.

\bibitem[{Dubey et~al.(2024)Dubey, Jauhri, Pandey, Kadian, Al-Dahle, Letman, Mathur, Schelten, Yang, Fan et~al.}]{dubey2024llama3}
Abhimanyu Dubey, Abhinav Jauhri, Abhinav Pandey, Abhishek Kadian, Ahmad Al-Dahle, Aiesha Letman, Akhil Mathur, Alan Schelten, Amy Yang, Angela Fan, et~al. 2024.
\newblock The llama 3 herd of models.
\newblock \emph{arXiv preprint arXiv:2407.21783}.

\bibitem[{Ethayarajh et~al.(2024)Ethayarajh, Xu, Muennighoff, Jurafsky, and Kiela}]{ethayarajh2024kto}
Kawin Ethayarajh, Winnie Xu, Niklas Muennighoff, Dan Jurafsky, and Douwe Kiela. 2024.
\newblock {KTO}: Model alignment as prospect theoretic optimization.
\newblock \emph{arXiv preprint arXiv:2402.01306}.

\bibitem[{Evans et~al.(2024)Evans, Parthasarathy, Merzic, and Henaff}]{evans2024batch_data_selection}
Talfan Evans, Nikhil Parthasarathy, Hamza Merzic, and Olivier~J Henaff. 2024.
\newblock Data curation via joint example selection further accelerates multimodal learning.
\newblock \emph{arXiv preprint arXiv:2406.17711}.

\bibitem[{Fang et~al.(2024)Fang, Lee, and Zhai}]{fang2024usinggpt4augmentunbalanced}
Luyang Fang, Gyeong-Geon Lee, and Xiaoming Zhai. 2024.
\newblock \href {https://arxiv.org/abs/2310.18365} {Using {GPT-4} to augment unbalanced data for automatic scoring}.
\newblock \emph{Preprint}, arXiv:2310.18365.

\bibitem[{Gao et~al.(2024)Gao, Hu, Ruan, Pu, and Wan}]{gao2024llmbasednlgevaluationcurrent}
Mingqi Gao, Xinyu Hu, Jie Ruan, Xiao Pu, and Xiaojun Wan. 2024.
\newblock \href {https://arxiv.org/abs/2402.01383} {{LLM}-based {NLG} evaluation: Current status and challenges}.
\newblock \emph{Preprint}, arXiv:2402.01383.

\bibitem[{Hu et~al.(2021)Hu, Wallis, Allen-Zhu, Li, Wang, Wang, Chen et~al.}]{hu2021loralowrankadaptationlarge}
Edward~J Hu, Phillip Wallis, Zeyuan Allen-Zhu, Yuanzhi Li, Shean Wang, Lu~Wang, Weizhu Chen, et~al. 2021.
\newblock Lora: Low-rank adaptation of large language models.
\newblock In \emph{International Conference on Learning Representations}.

\bibitem[{Jiang et~al.(2024)Jiang, Sablayrolles, Roux, Mensch, Savary, Bamford, Chaplot, Casas, Hanna, Bressand et~al.}]{jiang2024mixtral}
Albert~Q Jiang, Alexandre Sablayrolles, Antoine Roux, Arthur Mensch, Blanche Savary, Chris Bamford, Devendra~Singh Chaplot, Diego de~las Casas, Emma~Bou Hanna, Florian Bressand, et~al. 2024.
\newblock Mixtral of experts.
\newblock \emph{arXiv preprint arXiv:2401.04088}.

\bibitem[{Kaddour et~al.(2023)Kaddour, Harris, Mozes, Bradley, Raileanu, and McHardy}]{kaddour2023challenges}
Jean Kaddour, Joshua Harris, Maximilian Mozes, Herbie Bradley, Roberta Raileanu, and Robert McHardy. 2023.
\newblock Challenges and applications of large language models.
\newblock \emph{arXiv preprint arXiv:2307.10169}.

\bibitem[{Kaplan et~al.(2020)Kaplan, McCandlish, Henighan, Brown, Chess, Child, Gray, Radford, Wu, and Amodei}]{kaplan2020scaling}
Jared Kaplan, Sam McCandlish, Tom Henighan, Tom~B Brown, Benjamin Chess, Rewon Child, Scott Gray, Alec Radford, Jeffrey Wu, and Dario Amodei. 2020.
\newblock Scaling laws for neural language models.
\newblock \emph{arXiv preprint arXiv:2001.08361}.

\bibitem[{Kim et~al.(2024{\natexlab{a}})Kim, Shin, Cho, Jang, Longpre, Lee, Yun, Shin, Kim, Thorne et~al.}]{kim2024prometheusinducingfinegrainedevaluation}
Seungone Kim, Jamin Shin, Yejin Cho, Joel Jang, Shayne Longpre, Hwaran Lee, Sangdoo Yun, Seongjin Shin, Sungdong Kim, James Thorne, et~al. 2024{\natexlab{a}}.
\newblock Prometheus: Inducing fine-grained evaluation capability in language models.
\newblock In \emph{The Twelfth International Conference on Learning Representations}.

\bibitem[{Kim et~al.(2024{\natexlab{b}})Kim, Suk, Cho, Longpre, Kim, Yoon, Son, Cho, Shafayat, Baek, Park, Hwang, Jo, Cho, Shin, Lee, Oh, Lee, Ho, Joo, Ko, Lee, Chae, Shin, Jang, Ye, Lin, Welleck, Neubig, Lee, Lee, and Seo}]{kim2024biggenbenchprincipledbenchmark}
Seungone Kim, Juyoung Suk, Ji~Yong Cho, Shayne Longpre, Chaeeun Kim, Dongkeun Yoon, Guijin Son, Yejin Cho, Sheikh Shafayat, Jinheon Baek, Sue~Hyun Park, Hyeonbin Hwang, Jinkyung Jo, Hyowon Cho, Haebin Shin, Seongyun Lee, Hanseok Oh, Noah Lee, Namgyu Ho, Se~June Joo, Miyoung Ko, Yoonjoo Lee, Hyungjoo Chae, Jamin Shin, Joel Jang, Seonghyeon Ye, Bill~Yuchen Lin, Sean Welleck, Graham Neubig, Moontae Lee, Kyungjae Lee, and Minjoon Seo. 2024{\natexlab{b}}.
\newblock \href {https://arxiv.org/abs/2406.05761} {The {BiGGen Bench}: A principled benchmark for fine-grained evaluation of language models with language models}.
\newblock \emph{Preprint}, arXiv:2406.05761.

\bibitem[{Kim et~al.(2024{\natexlab{c}})Kim, Suk, Longpre, Lin, Shin, Welleck, Neubig, Lee, Lee, and Seo}]{kim2024prometheus2opensource}
Seungone Kim, Juyoung Suk, Shayne Longpre, Bill~Yuchen Lin, Jamin Shin, Sean Welleck, Graham Neubig, Moontae Lee, Kyungjae Lee, and Minjoon Seo. 2024{\natexlab{c}}.
\newblock \href {https://arxiv.org/abs/2405.01535} {Prometheus 2: An open source language model specialized in evaluating other language models}.
\newblock \emph{Preprint}, arXiv:2405.01535.

\bibitem[{Koutcheme et~al.(2024)Koutcheme, Dainese, Sarsa, Hellas, Leinonen, and Denny}]{10.1145/3649217.3653612}
Charles Koutcheme, Nicola Dainese, Sami Sarsa, Arto Hellas, Juho Leinonen, and Paul Denny. 2024.
\newblock \href {https://doi.org/10.1145/3649217.3653612} {Open source language models can provide feedback: Evaluating llms' ability to help students using gpt-4-as-a-judge}.
\newblock In \emph{Proceedings of the 2024 on Innovation and Technology in Computer Science Education V. 1}, ITiCSE 2024, page 52–58, New York, NY, USA. Association for Computing Machinery.

\bibitem[{Latif et~al.(2024)Latif, Fang, Ma, and Zhai}]{latif2024knowledgedistillationllmautomatic}
Ehsan Latif, Luyang Fang, Ping Ma, and Xiaoming Zhai. 2024.
\newblock \href {https://arxiv.org/abs/2312.15842} {Knowledge distillation of {LLM} for automatic scoring of science education assessments}.
\newblock \emph{Preprint}, arXiv:2312.15842.

\bibitem[{Latif and Zhai(2023)}]{latif2023finetuningchatgptautomaticscoring}
Ehsan Latif and Xiaoming Zhai. 2023.
\newblock \href {https://arxiv.org/abs/2310.10072} {Fine-tuning chatgpt for automatic scoring}.
\newblock \emph{Preprint}, arXiv:2310.10072.

\bibitem[{Lee et~al.(2023)Lee, Phatale, Mansoor, Mesnard, Ferret, Lu, Bishop, Hall, Carbune, Rastogi et~al.}]{lee2023rlaif}
Harrison Lee, Samrat Phatale, Hassan Mansoor, Thomas Mesnard, Johan Ferret, Kellie Lu, Colton Bishop, Ethan Hall, Victor Carbune, Abhinav Rastogi, et~al. 2023.
\newblock Rlaif: Scaling reinforcement learning from human feedback with ai feedback.
\newblock \emph{arXiv preprint arXiv:2309.00267}.

\bibitem[{Li et~al.(2024)Li, Xu, Shen, Xu, Gu, Lai, Tao, and Ma}]{li2024leveraginglargelanguagemodels}
Zhen Li, Xiaohan Xu, Tao Shen, Can Xu, Jia-Chen Gu, Yuxuan Lai, Chongyang Tao, and Shuai Ma. 2024.
\newblock \href {https://arxiv.org/abs/2401.07103} {Leveraging large language models for {NLG} evaluation: Advances and challenges}.
\newblock \emph{Preprint}, arXiv:2401.07103.

\bibitem[{Lin et~al.(2022)Lin, Hilton, and Evans}]{lin-etal-2022-truthfulqa}
Stephanie Lin, Jacob Hilton, and Owain Evans. 2022.
\newblock \href {https://doi.org/10.18653/v1/2022.acl-long.229} {{T}ruthful{QA}: Measuring how models mimic human falsehoods}.
\newblock In \emph{Proceedings of the 60th Annual Meeting of the Association for Computational Linguistics (Volume 1: Long Papers)}, pages 3214--3252, Dublin, Ireland. Association for Computational Linguistics.

\bibitem[{Mukherjee et~al.(2023)Mukherjee, Mitra, Jawahar, Agarwal, Palangi, and Awadallah}]{mukherjee2023orcaprogressivelearningcomplex}
Subhabrata Mukherjee, Arindam Mitra, Ganesh Jawahar, Sahaj Agarwal, Hamid Palangi, and Ahmed Awadallah. 2023.
\newblock \href {https://arxiv.org/abs/2306.02707} {Orca: Progressive learning from complex explanation traces of gpt-4}.
\newblock \emph{Preprint}, arXiv:2306.02707.

\bibitem[{Ouyang et~al.(2022)Ouyang, Wu, Jiang, Almeida, Wainwright, Mishkin, Zhang, Agarwal, Slama, Ray et~al.}]{ouyang2022instructGPT}
Long Ouyang, Jeffrey Wu, Xu~Jiang, Diogo Almeida, Carroll Wainwright, Pamela Mishkin, Chong Zhang, Sandhini Agarwal, Katarina Slama, Alex Ray, et~al. 2022.
\newblock Training language models to follow instructions with human feedback.
\newblock \emph{Advances in neural information processing systems}, 35:27730--27744.

\bibitem[{Penedo et~al.(2024)Penedo, Kydl{\'\i}{\v{c}}ek, Lozhkov, Mitchell, Raffel, Von~Werra, Wolf et~al.}]{penedo2024fineweb}
Guilherme Penedo, Hynek Kydl{\'\i}{\v{c}}ek, Anton Lozhkov, Margaret Mitchell, Colin Raffel, Leandro Von~Werra, Thomas Wolf, et~al. 2024.
\newblock The fineweb datasets: Decanting the web for the finest text data at scale.
\newblock \emph{arXiv preprint arXiv:2406.17557}.

\bibitem[{Rafailov et~al.(2024)Rafailov, Sharma, Mitchell, Manning, Ermon, and Finn}]{rafailov2024dpo}
Rafael Rafailov, Archit Sharma, Eric Mitchell, Christopher~D Manning, Stefano Ermon, and Chelsea Finn. 2024.
\newblock Direct preference optimization: Your language model is secretly a reward model.
\newblock \emph{Advances in Neural Information Processing Systems}, 36.

\bibitem[{Robinson(2014)}]{Robinson2014}
John Robinson. 2014.
\newblock \href {https://doi.org/10.1007/978-94-007-0753-5_1654} {\emph{Likert Scale}}, pages 3620--3621.
\newblock Springer Netherlands, Dordrecht.

\bibitem[{Wang et~al.(2024)Wang, Xu, Zhou, Xiong, and Joty}]{wang2024directjudgementpreferenceoptimization}
Peifeng Wang, Austin Xu, Yilun Zhou, Caiming Xiong, and Shafiq Joty. 2024.
\newblock \href {https://arxiv.org/abs/2409.14664} {Direct judgement preference optimization}.
\newblock \emph{Preprint}, arXiv:2409.14664.

\bibitem[{Wang et~al.(2023)Wang, Zhong, Li, Mi, Zeng, Huang, Shang, Jiang, and Liu}]{wang2023aligning}
Yufei Wang, Wanjun Zhong, Liangyou Li, Fei Mi, Xingshan Zeng, Wenyong Huang, Lifeng Shang, Xin Jiang, and Qun Liu. 2023.
\newblock Aligning large language models with human: A survey.
\newblock \emph{arXiv preprint arXiv:2307.12966}.

\bibitem[{Wei et~al.(2022)Wei, Wang, Schuurmans, Bosma, Xia, Chi, Le, Zhou et~al.}]{wei2023chainofthoughtpromptingelicitsreasoning}
Jason Wei, Xuezhi Wang, Dale Schuurmans, Maarten Bosma, Fei Xia, Ed~Chi, Quoc~V Le, Denny Zhou, et~al. 2022.
\newblock Chain-of-thought prompting elicits reasoning in large language models.
\newblock \emph{Advances in neural information processing systems}, 35:24824--24837.

\bibitem[{Zeng et~al.(2024)Zeng, Yu, Gao, Meng, Goyal, and Chen}]{zeng2024evaluatinglargelanguagemodels}
Zhiyuan Zeng, Jiatong Yu, Tianyu Gao, Yu~Meng, Tanya Goyal, and Danqi Chen. 2024.
\newblock \href {https://arxiv.org/abs/2310.07641} {Evaluating large language models at evaluating instruction following}.
\newblock \emph{Preprint}, arXiv:2310.07641.

\bibitem[{Zheng et~al.(2023)Zheng, Chiang, Sheng, Zhuang, Wu, Zhuang, Lin, Li, Li, Xing, Zhang, Gonzalez, and Stoica}]{zheng2023judgingllmasajudgemtbenchchatbot}
Lianmin Zheng, Wei-Lin Chiang, Ying Sheng, Siyuan Zhuang, Zhanghao Wu, Yonghao Zhuang, Zi~Lin, Zhuohan Li, Dacheng Li, Eric~P. Xing, Hao Zhang, Joseph~E. Gonzalez, and Ion Stoica. 2023.
\newblock \href {https://arxiv.org/abs/2306.05685} {Judging {LLM}-as-a-judge with {MT-Bench} and chatbot arena}.
\newblock \emph{Preprint}, arXiv:2306.05685.

\bibitem[{Zhu et~al.(2023)Zhu, Wang, and Wang}]{zhu2023judgelmfinetunedlargelanguage}
Lianghui Zhu, Xinggang Wang, and Xinlong Wang. 2023.
\newblock \href {https://arxiv.org/abs/2310.17631} {{JudgeLM}: Fine-tuned large language models are scalable judges}.
\newblock \emph{Preprint}, arXiv:2310.17631.

\end{thebibliography}

\section{Appendices}

% Use \verb|\appendix| before any appendix section to switch the section numbering over to letters. See Appendix~\ref{sec:appendix} for an example.

% \section{Bib\TeX{} Files}
% \label{sec:bibtex}

% Unicode cannot be used in Bib\TeX{} entries, and some ways of typing special characters can disrupt Bib\TeX's alphabetization. The recommended way of typing special characters is shown in Table~\ref{tab:accents}.

% Please ensure that Bib\TeX{} records contain DOIs or URLs when possible, and for all the ACL materials that you reference.
% Use the \verb|doi| field for DOIs and the \verb|url| field for URLs.
% If a Bib\TeX{} entry has a URL or DOI field, the paper title in the references section will appear as a hyperlink to the paper, using the hyperref \LaTeX{} package.

\appendix

\section{Preference Data Augmentation Prompt}
Feedback generation for each pair of questions and answers with LLM needs an input prompt.
Considering that we evaluate our approach on different datasets, we use adjusted input prompt templates for every dataset.
The following templates show the input prompts for the TruthfulQA dataset and the BigGen-Bench dataset.
These prompts are the output of the CoT prompt generator.
\label{sec:prompts}
\begin{tcolorbox}[colback=green!5!white, colframe=green!50!black, title= Judge Prompt (TruthfulQA), width=\columnwidth, sharp corners=south]
\label{prompt:truthful}
Given my answer to an instruction, your role is to provide specific and constructive feedback for me. You should find the best way for me to learn from your feedback and improve my performance.
You should consider multiple aspects of my answer, including helpfulness, truthfulness, honesty, and to what extent the answer follows instructions.

—

Instruction

${instruction}$

Answer

${answer}$

—

Please act as a teacher and provide specific and constructive feedback. Besides describing the weaknesses of the answer, you should also provide specific suggestions to guide me toward understanding how to improve. Please note, however, that your suggestions should help me better complete the instructions, but you should not introduce new requirements that are not mentioned in the instructions. Your feedback should focus on enhancing my ability to think critically and respond accurately. However, never explicitly provide the reference answer, nor do polite phrases be required. Only respond with concise feedback in chat style. Finally, score the overall quality of the answer from 1 to 10, where 1 is the worst and 10 is the best.
Format

Feedback

[Your feedback]

Overall Score: [1-10]

—

Feedback

\end{tcolorbox}

\begin{tcolorbox}[colback=green!5!white, colframe=green!50!black, title= Judge Prompt (BigGen-Bench), width=\columnwidth, sharp corners=south]
\label{prompt:bgb}
Task Description:

An instruction (might include an Input inside it), a response to evaluate, a reference answer that gets a score of 5, and a score rubric representing a evaluation criteria are given.

1. Write a detailed feedback that assess the quality of the response strictly based on the given score rubric, not evaluating in general.

2. After writing a feedback, write a score that is an integer between 1 and 5. You should refer to the score rubric.

3. You should refer to the score rubric to assign an integer score to response
4. The output format should look as follows: ""(write a feedback for criteria) [RESULT] (an integer number between 1 and 5)""

5. Please do not generate any other opening, closing, and explanations.

6. Be accurate in evaluating response and just give one score after in the end.

7. Don't get confused. You are conducting an absolute grading of another model's grading! For convenience, I will seperate the input and output of the other model's relative grading with ""@@@""s.

@@@

The instruction to evaluate:

[orig instruction]

@@@

Response to evaluate:

[orig response]

Reference Answer (Score 5):

[orig reference answer]

Score Rubrics:

[score rubric]

Feedback:

\end{tcolorbox}

\end{document}